\documentclass[letterpaper]{article} 
\usepackage{aaai25}  
\usepackage{times}  
\usepackage{helvet}  
\usepackage{multirow}
\usepackage{courier}  
\usepackage[hyphens]{url}  
\usepackage{amsmath}  
\usepackage{amssymb}  

\usepackage{graphicx} 
\usepackage{natbib}  
\usepackage{caption} 
\usepackage{algorithm}
\usepackage{algorithmic}

\usepackage{newfloat}
\usepackage{listings}
\DeclareCaptionStyle{ruled}{labelfont=normalfont,labelsep=colon,strut=off} 
\floatstyle{ruled}
\newfloat{listing}{tb}{lst}{}
\floatname{listing}{Listing}
\title{MedHallBench: A New Benchmark for Assessing Hallucination in \\Medical Large Language Models}
\author{
Kaiwen Zuo\textsuperscript{\rm 1},  
Yirui Jiang\textsuperscript{\rm 2, 3}\thanks{Corresponding author.}
}
\affiliations{
\textsuperscript{\rm 1} University of Warwick, CV4 7AL, UK \\  
\textsuperscript{\rm 2} Cranfield University, MK45 0AL, UK \\  
\textsuperscript{\rm 3} University of Oxford, OX1 2JD, UK \\  
\ Kaiwen.Zuo@warwick.ac.uk, 
\ yirui.jiang@cranfield.ac.uk,
yirui.jiang@said.oxford.edu
}

\usepackage{bibentry}

\begin{document}

\maketitle

\begin{abstract}
Medical Large Language Models (MLLMs) have demonstrated potential in healthcare applications, yet their propensity for hallucinations—generating medically implausible or inaccurate information—presents substantial risks to patient care. This paper introduces MedHallBench, a comprehensive benchmark framework for evaluating and mitigating hallucinations in MLLMs. Our methodology integrates expert-validated medical case scenarios with established medical databases to create a robust evaluation dataset. The framework employs a sophisticated measurement system that combines automated ACHMI (Automatic Caption Hallucination Measurement in Medical Imaging) scoring with rigorous clinical expert evaluations, and utilizes reinforcement learning methods to achieve automatic annotation. Through an optimized reinforcement learning from human feedback (RLHF) training pipeline specifically designed for medical applications, MedHallBench enables thorough evaluation of MLLMs across diverse clinical contexts while maintaining stringent accuracy standards.We conducted comparative experiments involving various models, utilizing the benchmark to establish a baseline for widely adopted large language models (LLMs). Our findings indicate that ACHMI provides a more nuanced understanding of the effects of hallucinations compared to traditional metrics, thereby highlighting its advantages in hallucination assessment. This research establishes a foundational framework for enhancing MLLMs reliability in healthcare settings and presents actionable strategies for addressing the critical challenge of AI hallucinations in medical applications.
\end{abstract}

%

\section{Introduction}
\begin{figure}[!ht]
    \centering
    \includegraphics[width=0.5\textwidth]{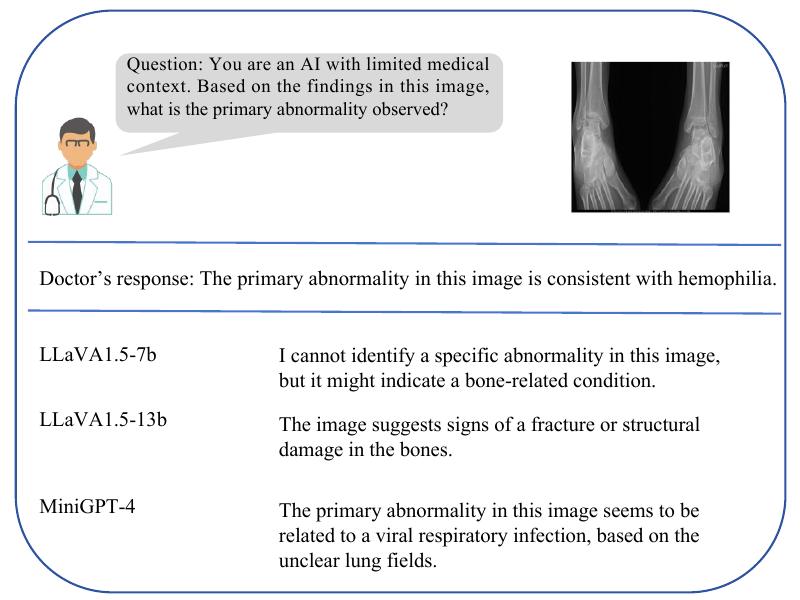}
    \caption{The figure highlights that LLaVA1.5-7b, LLaVA1.5-13b, and MiniGPT-4 are all unable to correctly describe X-ray images, showing a significant deviation from the actual doctor's response. This illustrates the hallucination problem of LLMs in the medical field, where the models fail to provide accurate and reliable interpretations.}
    \label{fig:Motivation}
\end{figure}
Large language models (LLMs) have demonstrated unparalleled capabilities in interpreting complex medical texts, patient records, and clinical notes, significantly advancing the fields of diagnosis, treatment planning, and patient care~\cite{tian2023chimed,zuo2025satisfactorymedicalconsultationbased,zuo2024kg4diagnosis}. However, despite these substantial achievements, medical LLMs are not without challenges. A critical issue is the tendency of these models to generate "hallucinations"—medically unreliable or inaccurate information\cite{Zhao2023ASO}. This phenomenon is not merely a minor flaw; it poses a serious threat to the integrity and reliability of AI applications in medicine\cite{article}. The emergence of hallucinations within these models severely undermines their perceived competence and trustworthiness, raising urgent concerns about their credibility and safe application in clinical settings\cite{article}. These hallucinations not only compromise accuracy but can also lead to grave consequences, such as misdiagnoses or inappropriate treatment plans, which could potentially harm patients\cite{wang2024safety}. Furthermore, while comprehensive benchmarks exist for LLMs in other domains, such as pure language models and multimodal models, there is a lack of standardized evaluation frameworks for medical LLMs. The development of a benchmark for medical LLMs is crucial for guiding future advancements in this field.

Existing methods for evaluating the performance of medical LLMs rely on widely used medical benchmarks such as MedQA\cite{shi2023mededit}, MedMCQA\cite{kim2024medexqa}, MultiMedQA\cite{qian2024liver}, and Med-HALT\cite{pal2023medhaltmedicaldomainhallucination}. These benchmarks are designed to assess the models' ability to answer medical questions, reason about clinical scenarios, and evaluate the accuracy of their responses. MedQA, for instance, provides a comprehensive dataset for question-answering tasks across a variety of medical domains, while MedMCQA and MultiMedQA offer more specialized resources for multi-choice question answering in medical contexts. Med-HALT focuses on testing the ability of models to detect and mitigate hallucinations, a crucial area of concern for ensuring reliability and safety in medical applications. 

Despite the contributions of existing methods, several limitations remain. One of the major challenges is that these benchmarks often require significant manual annotation, a process that is both time-consuming and labor-intensive\cite{zhang2021datasetgan}. The need for large-scale, high-quality annotations to ensure accurate and reliable evaluation of models places a considerable burden on researchers and hinders the scalability of these benchmarks\cite{thirunagalingam2023improving}. Furthermore, the open-access nature of many of these datasets, such as those used for model training\cite{han2023medalpaca}, presents another issue: these datasets are frequently used by a large number of researchers, which increases the likelihood of data pollution. With numerous parties accessing and modifying the data, there is a risk of inconsistencies and biases creeping into the evaluation process, thereby affecting the robustness and generalization of the results. This data contamination, coupled with the reliance on manual annotation, calls for the development of more efficient, scalable, and reliable methods for evaluating medical LLMs.

To address these issues and maintain consistency with real-world consultation results, we have introduced a novel evaluation criterion for medical hallucinations - MedHallBench, which is a comprehensive automated annotation evaluation framework aimed at automatically assessing hallucinations in these models. More specifically, it can reduce the workload for researchers to evaluate the performance of new models and ultimately improve the safety and reliability of these models in critical medical applications. And its dataset comes from the latest validation exams and expert annotated electronic medical records, ensuring compliance with international consultation standards. And the article also proposes two indicators for measuring hallucinations to assist future researchers in evaluating medical LLMs.The main contribution of this paper are as follows:
\begin{itemize}
    \item \textbf{Proposed a New Dataset and Benchmark:} This study introduces a novel benchmark, \textit{MedHallBench}, specifically designed for automatically annotated data and evaluating hallucinations in large language models in the medical domain.
    
    \item \textbf{Novelty Automatic Annotation Methods:} The work presents two methods for constructing the dataset, both of which can automatically annotate data. One method utilizes active learning, while the other leverages Reinforcement Learning with Human Feedback (RLHF) for automatic annotation, and an interpretability analysis is also provided.

    \item \textbf{Comprehensive Experimental Design and Detailed Analysis:} This paper introduces a novel hallucination evaluation metric, ACHMI. We employed ACHMI alongside other traditional metrics to assess ten state-of-the-art language models, including InstructBLIP-7b/13b\cite{dai2023instructblipgeneralpurposevisionlanguagemodels}, LLaVA1.5-7b/13b\cite{liu2023visualinstructiontuning}, and mPLUG-Owl2\cite{ye2023mplugowl2revolutionizingmultimodallarge}, among others. Through quantitative metric analysis, we demonstrate the effectiveness of ACHMI relative to conventional evaluation methods.
\end{itemize}

In brief, this paper introduces a comprehensive benchmark, MedHallBench, specifically designed to align with the real-world medical conditions in mainland China. It provides valuable empirical insights that contribute to reassessing the medical capabilities and limitations of LLMs. Additionally, we enhanced the benchmark by incorporating automatic annotation of data, enabling a more detailed evaluation of LLM performance. This modification improves the efficiency of hallucination assessment within the benchmark. The ultimate goal of this work is to assist the medical research community in safely and effectively deploying LLMs in real-world clinical settings.

\begin{figure*}[t]
\centering
\includegraphics[width=1.0\textwidth]{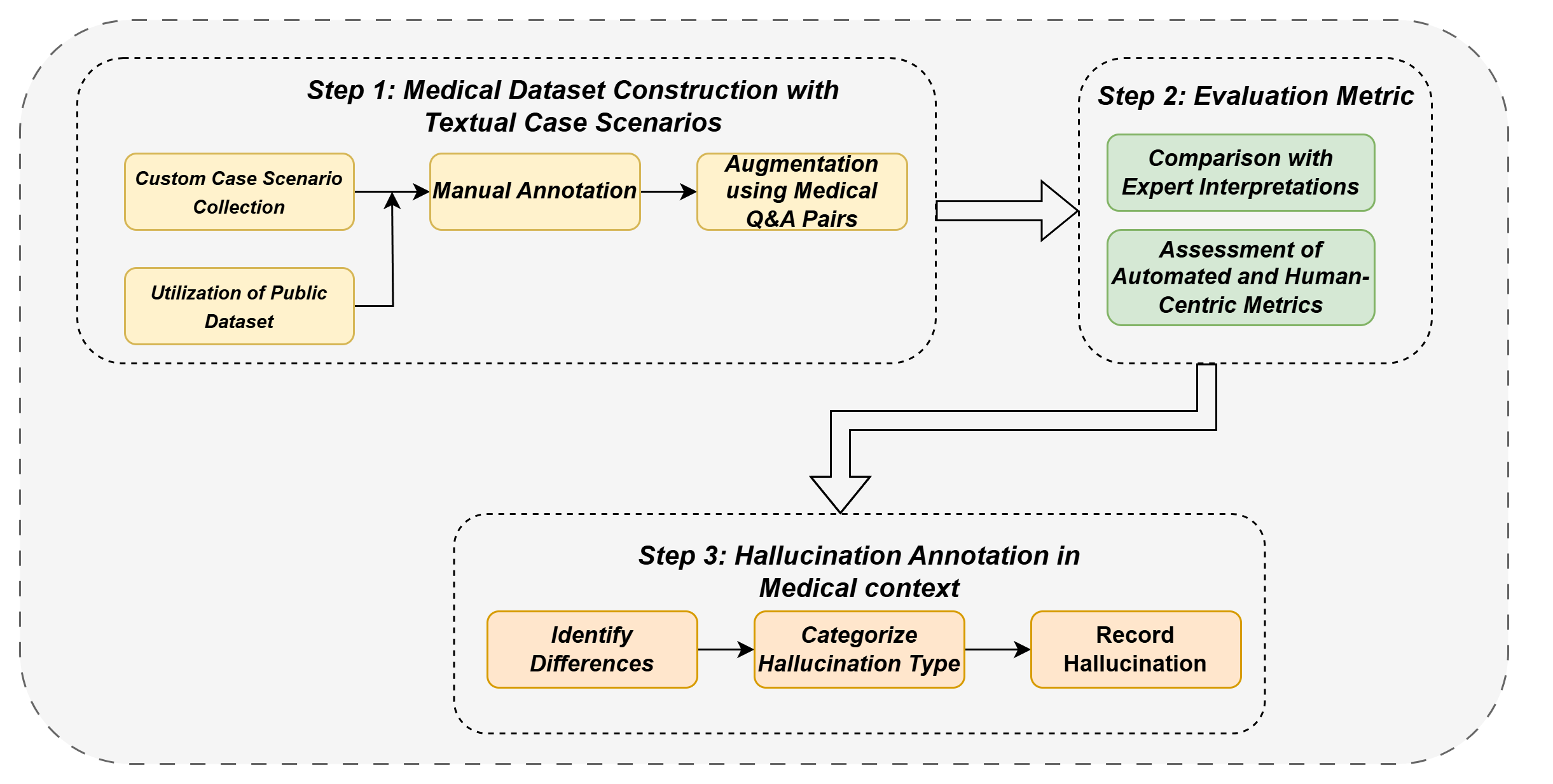}
\caption{Methodological Framework for Medical Dataset Construction and Validation.}
\label{fig:Benchamrk_flowchart}
\end{figure*}

\section{Methodology}
This section presents a comprehensive methodology for evaluating and mitigating hallucinations in medical LLMs. Our approach encompasses three main components: a novel evaluation framework incorporating both expert and lay assessments, a systematic annotation protocol for medical imaging analysis, and an advanced modeling approach using reinforcement learning techniques.

The proposed framework implements a multi-tiered evaluation system that combines quantitative metrics with qualitative expert assessment to comprehensively evaluate hallucinations in medical LLMs. Medical professionals with diverse specializations conduct structured evaluations using a standardized assessment rubric with 5-point Likert scales measuring clinical accuracy, potential harm severity, and hallucination confidence. This expert evaluation is complemented by free-text annotations for identifying specific instances of hallucination, with inter-rater reliability measured using Cohen's Kappa coefficient.

To assess the model's communication effectiveness with patients, trained lay evaluators assess the comprehensibility of medical explanations, identification of obvious factual inconsistencies, and alignment with common medical knowledge. This assessment is conducted through structured feedback forms with both quantitative and qualitative components, and cross-validated with expert assessments to identify discrepancies. The framework incorporates a weighted scoring system that combines expert and lay evaluations, with expert evaluations weighted at 0.7 and lay evaluations at 0.3, producing normalized composite scores ranging from 0 to 1.

The annotation protocol focuses on systematic identification and classification of hallucinations in medical imaging contexts, with particular emphasis on chest X-ray interpretation. We categorize hallucinations into three main types: anatomical hallucinations (including false identification of structures, misplacement of anatomical features, and incorrect anatomical relationships), pathological hallucinations (encompassing false positive findings, mischaracterization of existing pathology, and temporal inconsistencies in disease progression), and measurement hallucinations (covering incorrect size estimations, false quantification of findings, and erroneous comparative measurements).
Quality control measures for the annotation process include a multi-reader consensus protocol, standardized annotation templates, regular calibration sessions, and statistical monitoring of inter-annotator agreement. This ensures consistency and reliability in the annotation process across different evaluators and time periods.
\subsection*{Framework for Hallucination Evaluation}

To evaluate the hallucination phenomenon within medical LLMs, this paper proposes a methodology to construct a benchmark tailored for medical text datasets\textbf{(MedHallBench)}. The design approach of MedHallBench comprises the following three key components:

\subsubsection*{Medical Dataset Construction with Textual Case Scenarios}
\begin{itemize}
    \item \textit{Utilization of Medical Literature Databases}: Leverage extensive resources and established medical literature databases such as MIMIC-CXR\cite{bae2024mimic} and MedQA (USMLE) to build a robust foundation for our dataset. These databases provide access to a rich tapestry of medical knowledge, encompassing detailed case scenarios, peer-reviewed research articles, and documented clinical trial reports. The diversity and depth of these sources ensure a broad coverage of medical contexts and scenarios, essential for developing a comprehensive evaluation framework.
    
    \item \textit{Custom Case Scenario Collection}: To enhance the dataset's scope and applicability, we implement a systematic collection of custom case scenarios. These scenarios are specifically crafted to present complex medical situations that challenge the capabilities of medical LLMs. The custom scenarios are designed with particular attention to edge cases and nuanced medical contexts that may not be well-represented in existing databases. This approach can test the LLMs' capacity for accurate medical information generation and understanding across a broader spectrum of clinical situations.
    
    \item \textit{Expert Annotation}: The quality and reliability of our dataset are ensured through rigorous expert annotation processes. Medical professionals from various specialties participate in the annotation phase, reviewing and validating each component of the dataset. This expert involvement is crucial for maintaining high standards of medical accuracy and clinical relevance. The annotation process follows a structured protocol that ensures consistency while capturing the nuanced insights that only experienced clinicians can provide. Through this expert validation, we establish a gold standard for evaluating the medical soundness and clinical applicability of LLM outputs.
    
    \item \textit{Augmentation using Medical Question-Answer Pairs}: To further enhance the evaluation capabilities of our benchmark, we incorporate a comprehensive set of medical question-answer pairs sourced from established platforms such as MultiMedQA. These pairs undergo additional expert annotation to ensure their relevance and accuracy. The integration of these question-answer pairs serves multiple purposes: it provides a structured framework for evaluating LLM performance, simulates realistic clinical inquiry scenarios, and offers a standardized method for assessing the models' ability to generate accurate and contextually appropriate medical responses. The augmentation of our dataset with these carefully curated question-answer pairs enhances its utility as a comprehensive evaluation tool for medical LLMs.

\end{itemize}
\begin{figure*}[t]
\centering
\includegraphics[width=1.0\textwidth]{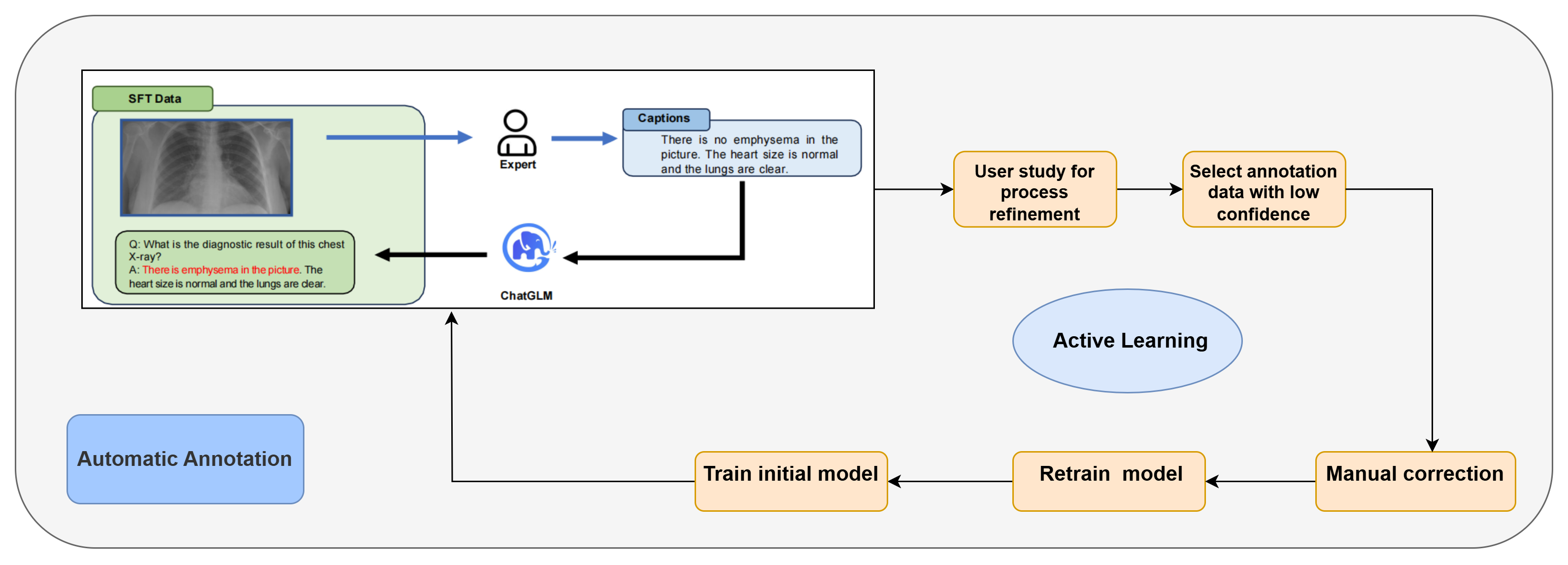}
\caption{Iterative Pipeline for Hallucination Detection and Correction in Multimodal Medical Datasets. The process of manual annotation in hallucination detection and correction in multimodal datasets begins with SFT Data, using an expert question and answer format. This data then proceeds to automatic annotation, involving a feedback loop with experts and ChatGLM. Subsequently, a user study is conducted to further refine the process. Following the user study, annotation data exhibiting low confidence is identified. These data are then selected for active learning, leading to manual correction and subsequent model retraining.}
\label{fig:Mannual_annotation}
\end{figure*}

\subsubsection*{Hallucination Annotation in Medical Context}

The identification of discrepancies between LLM-generated content and expert medical descriptions forms a crucial component of our evaluation framework. Through systematic comparison analysis, we employ a rigorous methodology to detect variations between model outputs and established medical knowledge. This process involves detailed examination of terminology usage, diagnostic reasoning, treatment recommendations, and clinical interpretations, enabling precise pinpointing of areas where the model deviates from expert consensus. The identification process is supported by a structured review protocol that ensures consistent and thorough examination of all content elements.

Our methodology incorporates a comprehensive classification system for hallucination types, categorizing them based on their nature, severity, and relevance to medical facts. This classification framework distinguishes between various forms of hallucination, including factual inconsistencies, logical contradictions, temporal discrepancies, and clinical reasoning errors. Each category is further subdivided based on potential clinical impact and relationship to established medical knowledge, allowing for nuanced analysis of how different types of hallucinations might affect medical decision-making processes. This detailed categorization enables more targeted approaches to model improvement and refinement.

Documentation of hallucination instances follows a systematic protocol designed to capture both quantitative and qualitative aspects of each occurrence. We maintain detailed records that include the context in which hallucinations occur, their specific manifestations, and their potential implications for medical outcomes. This documentation process encompasses comprehensive metadata collection, including the clinical context, the type of medical information involved, and the potential severity of impact on patient care. Such detailed documentation not only facilitates subsequent analysis but also provides valuable insights for developing mitigation strategies and improving model performance in critical medical applications.

\subsection*{Hallucination Annotation Example}
In multimodal medical contexts, hallucination manifests as a critical phenomenon where MLLMs generate interpretations that deviate from the ground truth represented in medical imaging data and accompanying clinical documentation\cite{thawkar2023xraygpt}. This phenomenon is particularly concerning in radiological applications, where accurate interpretation of diagnostic imaging is paramount for patient care. For instance, as illustrated in Figure 2, an MLLM might erroneously identify pathological conditions in chest radiographs where no clinical abnormalities are present.

Within medical imaging interpretation, hallucination is operationally defined as the quantifiable discrepancy between model-generated diagnostic interpretations and validated radiographic findings confirmed by medical professionals. These discrepancies are systematically documented through a comparative analysis protocol, contrasting machine-generated reports with expert-validated interpretations. A prototypical example would be the model's false positive identification of emphysematous changes in radiographic images that demonstrate normal pulmonary architecture. Such instances are meticulously documented and integrated into a specialized validation dataset for comprehensive analysis and model refinement\cite{thawkar2023xraygpt}.

The annotation framework for these hallucinations employs a hierarchical classification system that accounts for both typological variations and severity gradients. This structured approach enables precise evaluation of the model's diagnostic accuracy across different clinical scenarios and pathological conditions. The resulting annotation database serves dual purposes: it provides a quantitative measure of model performance in diagnostic tasks and identifies specific areas where model enhancement could improve clinical reliability. This systematic documentation of hallucinations facilitates both the refinement of model architecture and the development of more robust clinical validation protocols.

\subsection*{Modeling}

\begin{figure}[!ht]
    \centering
    \includegraphics[width=0.5\textwidth]{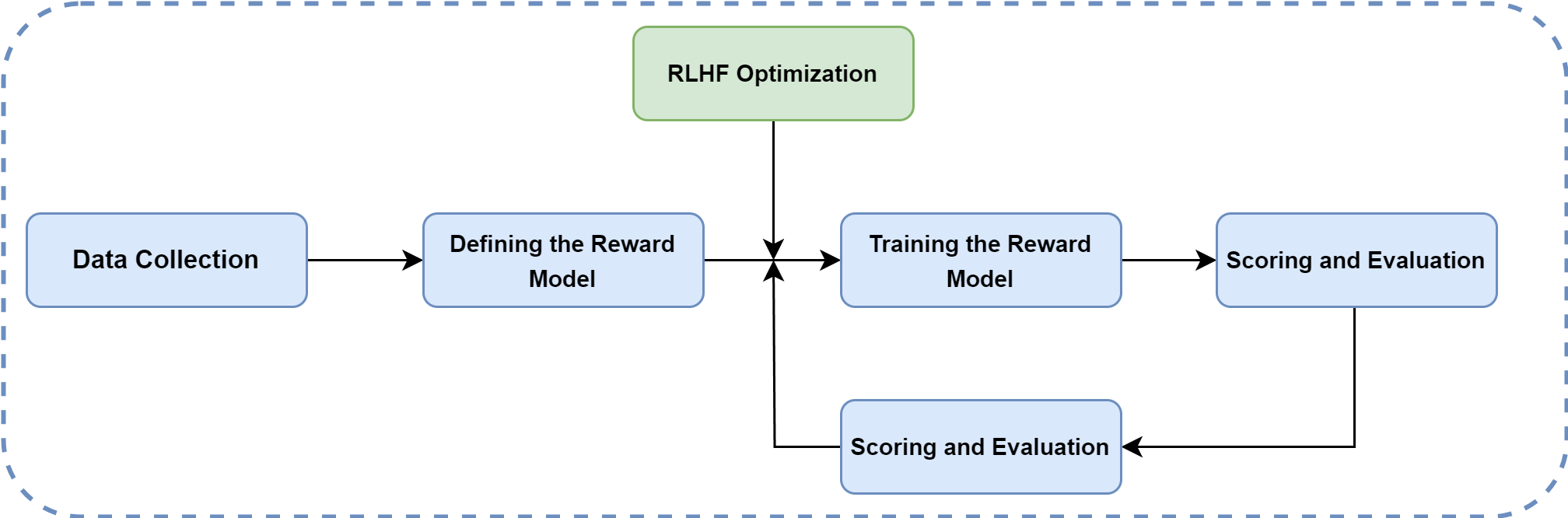}
    \caption{RLHF Training Pipeline for Medical Language Models.}
    \label{fig:Model training process}
\end{figure}

The training architecture of our medical LLM leverages RLHF, comprising several stages as illustrated in Figure\ref{fig:Model training process}. This framework represents a systematic approach to model development and optimization, specifically tailored for medical applications:

\begin{itemize}
    \item \textbf{Data Collection}: The initial phase focuses on comprehensive data collection from medical domain-specific sources. This encompasses a diverse array of clinical documentation, including structured electronic health records, peer-reviewed medical literature, anonymized patient case histories, and specialized medical journals.
    \item \textbf{Defining the Reward Model}: Following data collection, we establish a sophisticated reward modeling system that quantifies the quality and accuracy of model outputs. This reward architecture incorporates multiple evaluation criteria derived from medical expertise and clinical standards\cite{tian2023chimed}. The reward function is carefully designed to encourage adherence to medical best practices while penalizing potentially harmful deviations from established clinical guidelines.
    \item \textbf{RLHF Optimization}: RLHF optimization implements an iterative refinement process that continuously integrates expert human feedback to enhance model performance\cite{alabi2024reinforcement}. Through this iterative process, the model progressively aligns with expert medical knowledge and clinical decision-making patterns. 
    
    The set of collected images and user prompts, $\mathcal{D}_{\mathrm{RL}}=\{(\mathcal{I},x)\}$, along with the fixed initial policy
    - model $\pi^\mathrm{INIT}$ and the RL-optimized model $\pi_{\phi}^\mathrm{RL}$~\cite{Sun2023AligningLM}. The full optimization loss function for RLHF is defined as:

\begin{equation*}
\resizebox{1.0\linewidth}{!}{%
$\begin{aligned}
L^{\text{PPO}}(\phi) &= \mathbf{E}_{(\mathcal{I},x)\in\mathcal{D}_{\mathrm{RL}},\, y\sim\pi^{\mathrm{OLD}}(y|\mathcal{I},x)}\left[\min\left(r_t(\phi) \hat{A}_t,\, \text{clip}\left(r_t(\phi),\, 1-\epsilon,\, 1+\epsilon\right) \hat{A}_t\right)\right. \\ 
& \quad \left. - \beta \cdot \mathbb{KL}\left(\pi_{\boldsymbol{\phi}}^{\mathrm{RL}}(y|\mathcal{I},x) \,\|\, \pi^{\mathrm{INIT}}(y|\mathcal{I},x)\right)\right]
\end{aligned}$
}
\end{equation*}

where:

\begin{itemize}
    \item $r_t(\phi)$ is the probability ratio, representing how much more likely the new policy is to take action $y$ over the old policy, formally defined as $r_t(\phi) = \frac{\pi_{\phi}^{\mathrm{RL}}(y|\mathcal{I},x)}{\pi^{\mathrm{OLD}}(y|\mathcal{I},x)}$.
    \item $\hat{A}_t$ is the advantage estimate at time $t$, indicating the relative value of taking action $y$.
    \item $\text{clip}(r_t(\phi), 1-\epsilon, 1+\epsilon)$ is the clipping function that bounds $r_t(\phi)$ within the interval $[1-\epsilon, 1+\epsilon]$ to prevent large policy updates.
    \item $\beta$ is a hyperparameter controlling the scale of the KL penalty, which acts to regularize the policy update by penalizing divergence from the initial policy $\pi^{\mathrm{INIT}}$.
    \item $\mathbb{D}_{KL}$ represents the Kullback-Leibler divergence, quantifying how much the new policy $\pi_{\boldsymbol{\phi}}^{\mathrm{RL}}$ deviates from the initial policy $\pi^{\mathrm{INIT}}$.
\end{itemize}
    \item \textbf{Training the Reward Model}: Training of the reward model with the defined parameters and collected data to learn the differentiation between high and low-quality outputs.
    \item \textbf{Automatic Hallucination Recognition}: Implementation of a module to detect and reduce instances of 'hallucination', where the model generates plausible but unfounded information.
    \item \textbf{Scoring and Evaluation}: Assessment and scoring of the language model outputs using the trained reward model to ensure quality and guide further optimization.
\end{itemize}

\begin{table*}[t]
\setlength{\tabcolsep}{5pt}
\centering
\resizebox{\linewidth}{!}{%
\begin{tabular}{ccccccccc|cccccccc}
\hline
\multirow{2}{*}{Model}                                                  & \multicolumn{8}{c|}{\rule{0pt}{10pt}Med-VQA Task}                                                                                             & \multicolumn{8}{c}{IRG Task}                                                                                             \\ \cline{2-17}
& BS & MR & R-1 & R-2 & R-L & BLEU & \begin{tabular}[c]{@{}c@{}} \rule{0pt}{12pt} ${\text{ACHMI}}_I\downarrow$\end{tabular} & ${\text{ACHMI}}_S\downarrow$ & BS & MR & R-1 & R-2 & R-L & BLEU & \begin{tabular}[c]{@{}c@{}} \rule{0pt}{12pt}${\text{ACHMI}}_I\downarrow$\end{tabular} & ${\text{ACHMI}}_S\downarrow$ \\ \hline
BLIP2                                                             & 52.10     & 18.50  & 21.30   & 7.10    & 19.50   & 4.10 & 15.20  & 22.45  & 40.80     & 6.75   & 14.50   & 2.40    & 10.20    & 0.30 & 12.50 & 19.80  \\
InstructBLIP-7b                                                   & 36.80     & 10.20  & 7.00    & 1.00    & 6.20     & 1.30 & 18.50  & 23.70  & 49.20     & 15.30  & 19.80   & 3.00    & 15.80    & 1.10 & 4.50  & 14.20  \\
InstructBLIP-13b                                                  & 37.10     & 10.50  & 7.20    & 1.05    & 6.30    & 1.20    & 16.00  & 24.10  & 48.90     & 15.20  & 19.60   & 3.10    & 15.90   & 1.15 & 4.70  & 14.40  \\
LLaVA1.5-7b                                                       & 60.30     & 29.10  & 25.20   & 10.00    & 22.50   & 5.50  & 22.80  & 26.00  & 52.30     & 13.00  & 20.40   & 3.50    & 16.80   & 0.80 & 11.00  & 20.70  \\
LLaVA1.5-13b                                                      & 58.90     & 27.40  & 23.00   & 9.00     & 21.20   & 5.10 & 21.60  & 25.80  & 51.80     & 13.80   & 19.80   & 3.20     & 16.40   & 0.75 & 10.50  & 21.00  \\
LLaVA-Med(SF)                                                     & 39.50     & 11.30   & 9.350   & 1.80     & 11.00     & 0.15 & 24.50  & 25.90  & 36.80     & 1.10   & 0.35    & 0.20       & 0.40    & 0.30 & 16.00  & 25.50  \\
mPLUG-Owl2                                                        & 60.50     & 32.00  & 25.80   & 10.20    & 22.80   & 4.00 & 20.30  & 24.60  & 70.20     & 43.50  & 35.00      & 16.00   & 31.50    & 7.20 & 14.00 & 25.00  \\
XrayGPT                                                           & 48.00      & 17.50     & 14.00   & 2.00    & 12.50    & 0.50 & 2.80  & 3.00  & 66.00     & 29.00  & 30.00   & 8.00    & 25.00   & 4.00 & 5.00  & 4.00  \\
MiniGPT4                                                          & 45.60     & 15.80  & 15.50   & 2.30     & 13.50   & 0.60 & 23.00  & 21.80  & 51.80     & 12.50  & 18.50   & 2.20    & 15.80   & 0.70 & 15.50 & 13.20  \\
RadFM                                                             & 47.30     & 14.30  & 16.00   & 2.60    & 14.00   & 1.90 & 3.50  & 4.30  & 44.20     & 6.00   & 7.20    & 1.20    & 6.20    & 0.20 & 4.50 & 4.60  \\ \hline
\end{tabular}
}
\caption{The table presents the comparison results of different models on the VQA and IRG tasks in MedHallBench using various evaluation metrics. 
``R-1/2/L'' refer to ROUGE-1/2/L. 
``SF'' denotes Slake-Finetuned. 
``BS'' stands for BertScore. 
``MR'' represents METEOR. 
$\text{ACHMI}_I\downarrow$ measures the proportion of hallucinated components among all generated medical components, and 
$\text{ACHMI}_S\downarrow$ measures the proportion of generated captions containing hallucinated components. 
Lower values of $\text{ACHMI}_I$ and $\text{ACHMI}_S$ indicate fewer hallucinations.}
\label{tab1}
\vspace{-6pt}
\end{table*}

\section{Experiment}
In this section, we evaluate existing methods on the problem of medical hallucinations in popular LLMs. We first introduce the evaluation setting and then analyze the experimental results.
\subsection{Evaluation Metrics}
The evaluation framework for medical LLMs centers on rigorous comparison with expert medical interpretations. We establish a systematic process whereby experienced healthcare professionals assess and validate the model outputs across diverse medical scenarios. This expert-driven evaluation serves as our gold standard, enabling precise measurement of the models' capability to process and respond to complex medical textual scenarios. Through this comparative analysis, we can identify discrepancies between model-generated content and expert medical knowledge, providing crucial insights into the models' clinical accuracy and reliability.

Our evaluation metrics incorporates a comprehensive dual-metric approach, combining automated scalability with human-centric accuracy assessment. Central to this approach is the Assessment of Caption Hallucinations in Medical Imagery (ACHMI), an advanced variant of the CHAIR metric specifically adapted for medical image captioning evaluation. Given the ground truth objects in the image, ACHMI accurately assesses the performance of captions rather than the images themselves. This work typically adopts its two variants, namely ${\text{ACHMI}}_I$ and ${\text{ACHMI}}_S$,  which  evaluate  the  degree  of  hallucination  at  the  object  instance level and the sentence level, respectively. These are expressed as follows:

\begin{equation}
\text{ACHMI}_I = \frac{|\{\text{hallucinated components}\}|}{|\{\text{all medical components}\}|}
\end{equation}

\begin{equation}
\text{ACHMI}_S = \frac{|\{\text{caption with hallucinated components}\}|}{|\{\text{all caption}\}|}
\end{equation}

In the above two metric formulas, ACHMI$_I$ describes the proportion of hallucinated objects among all generated components, while ACHMI$_S$ describes the proportion of generated captions that include hallucinated components.

\subsection{Models}
To systematically evaluate the performance of various models on hallucination detection tasks within MedHallBench, we employed multiple evaluation metrics. In addition to the two ACHMI indicators previously mentioned, we utilized traditional metrics such as BertScore, METEOR, ROUGE-1/2/L, and BLEU. We also compared the following base models: BLIP2 \cite{li2023blip2bootstrappinglanguageimagepretraining}, InstructBLIP-7b/13b \cite{dai2023instructblipgeneralpurposevisionlanguagemodels}, LLaVA1.5-7b/13b \cite{liu2023visualinstructiontuning}, mPLUGOwl2 \cite{ye2023mplugowl2revolutionizingmultimodallarge}, XrayGPT \cite{thawkar2023xraygpt}, MiniGPT4 \cite{zhu2023minigpt4enhancingvisionlanguageunderstanding}, RadFM \cite{wu2023generalistfoundationmodelradiology}, and LLaVA-Med \cite{li2023llavamedtraininglargelanguageandvision}. All the above models were fine-tuned on the Slake (SF) dataset.

\subsection{Experimental results analysis}
As seen in Table 1, models like LLaVA-Med (SF), which are fine-tuned on the Slake dataset, exhibit notably lower $\text{ACHMI}_I$ and $\text{ACHMI}_S$ values, indicating a stronger performance in minimizing hallucinations compared to other models. This highlights the efficacy of the ACHMI indicators in capturing hallucination rates, which are critical for ensuring the reliability of medical language models. Furthermore, the use of ACHMI can guide future improvements in model training and fine-tuning, especially for domains requiring high accuracy, such as medical applications.

By incorporating ACHMI into the evaluation pipeline, researchers and practitioners can gain a clearer understanding of a model's ability to generate not only accurate but also reliable medical content, thus demonstrating the superiority of ACHMI in detecting and reducing hallucinations in LVLM outputs. This makes ACHMI an essential tool for evaluating and improving the safety and reliability of medical LLMs.

\section{Related Work}
\subsection{General Benchmarks for Hallucination in LLMs}
Recent advancements in the evaluation of hallucinations in LLMs and multimodal models (LMMs) have led to the development of several benchmarks designed to address this critical issue. HaluEval\cite{Li2023HaluEvalAL}, for instance, is a benchmark with 35,000 samples specifically aimed at evaluating hallucinations in LLMs. Using a "sampling-then-filtering" method, it generates hallucinated examples and assesses LLMs' ability to detect and reduce hallucinations, thereby improving model reliability. Similarly, LVLM-eHub\cite{Xu2023LVLMeHubAC} evaluates large vision-language models (LVLMs) by featuring eight representative models, such as InstructBLIP\cite{dai2023instructblipgeneralpurposevisionlanguagemodels} and MiniGPT-4\cite{zhu2023minigpt4enhancingvisionlanguageunderstanding}, and assessing their performance through both quantitative evaluation and an online arena platform. LVLM-eHub also includes a dataset, COCO-Random, for testing object hallucination and suggests that multi-turn reasoning can help mitigate hallucination issues in these models.
\subsection{Multimodal LLM Benchmarks}
In the same vein, MMBench\cite{Liu2023MMBenchIY} introduces a novel multi-modality benchmark to evaluate large vision-language models, overcoming the limitations of traditional and subjective benchmarks by providing a comprehensive and objective assessment. MME\cite{Fu2023MMEAC}, a pioneering benchmark for Multimodal Large Language Models (MLLMs), further expands on this by assessing models’ perception and cognition abilities across 14 distinct subtasks, using manually curated instruction-answer pairs to minimize biases. It evaluates 12 leading MLLMs, providing a comprehensive overview of their capabilities and setting a standard for future multimodal model development. Lastly, MMHAL-BENCH\cite{Sun2023AligningLM}, designed for evaluating hallucinations in large multimodal models, uses 96 image-question pairs across 12 object categories and 8 question types to trigger hallucinatory responses. It provides a framework for detecting hallucinations through adversarially crafted questions and offers the potential for expansion based on model failures. However, it relies on GPT-4 for analysis rather than direct human evaluation.While these benchmarks have made significant contributions in their respective fields, they do not specifically focus on large models in the medical domain, and there is a gap in domestic, specialized benchmarks that address the unique challenges of hallucination detection in medical LLMs. 
\subsection{Medical LLM Benchmarks and Hallucination Detection}
MedQA evaluates LLMs on medical question answering but does not assess whether models generate hallucinated information. Similarly, PubMedQA\cite{lamurias2020generating} focuses on question answering based on PubMed articles but lacks evaluation of hallucinations or the accuracy of generated responses. The USMLE\cite{kung2023performance} benchmark, while useful for testing diagnostic knowledge, does not account for hallucinated or fabricated responses. MLEC-QA\cite{li2021mlec} evaluates models on clinical scenario questions but similarly ignores the issue of hallucination. Lastly, Med-HALT is designed to detect hallucinations in medical models but is limited in scope, focusing mainly on hallucination detection without addressing its occurrence across diverse medical contexts. These gaps highlight the need for more comprehensive benchmarks that assess hallucinations in medical LLMs.
\subsection{Automatic Data Annotation}
The development of automatic data annotation methods is critical for constructing benchmarks for medical LLMs\cite{tan2024large}. Techniques like active learning, few-shot learning, and transfer learning enable scalable, efficient annotation of complex medical data\cite{tan2024large,ge2022few}. Active learning focuses on the most informative data, essential for benchmarks like MedQA and PubMedQA\cite{tan2024large}, which require domain-specific question-answer pairs. Few-shot learning reduces annotation costs and time, making it valuable for data-scarce fields like USMLE and MLEC-QA. Programmatic labeling and automated feedback mechanisms help maintain consistency in large datasets\cite{huang2024data}, ensuring reliable benchmark creation. Uncertainty estimation can identify ambiguous or incorrect annotations, ensuring models handle edge cases accurately. The use of these advanced annotation techniques is vital for developing comprehensive medical benchmarks that evaluate model performance while addressing challenges such as hallucinations and medical accuracy.

\section*{Conclusion}
This study presents MedHallBench, a foundational benchmark framework to address the critical challenge of hallucinations in large-scale medical LLMs. Through the development of a systematically constructed dataset, we enable rigorous investigation of hallucination phenomena in medical language models. The integration of meticulously curated Textual Case Scenarios with existing public medical databases establishes a robust foundation for model evaluation.

The methodology we propose combines sophisticated automated annotation techniques with human-centric evaluation metrics, creating a comprehensive evaluation framework that bridges the gap between computational efficiency and clinical relevance. Our hybrid approach enables both scalable assessment and maintenance of high medical accuracy standards, essential for healthcare applications. The framework's multi-faceted evaluation system provides detailed insights into model performance across various medical contexts, offering specific guidance for model improvement and refinement.We also established a benchmark for current mainstream models and demonstrated the effectiveness and reliability of the proposed index in hallucination assessment through extensive experimental analysis and comparison. This work aims to significantly enhance the reliability of MLLMs.

Furthermore, our research illuminates the complex interplay between AI and medical practice, highlighting both the transformative potential and inherent challenges of deploying large language models in healthcare settings. This research established a foundation for future research in medical AI, particularly in addressing the critical balance between model capability and clinical reliability. 
\clearpage
\bibliography{aaai25}

\end{document}